\documentclass{article} 
\usepackage[final]{colm2026_conference}

\usepackage{microtype}
\usepackage{hyperref}
\usepackage{url}
\usepackage{booktabs}
\usepackage{graphicx}

\usepackage{caption}
\usepackage{xcolor}
\usepackage{tcolorbox}
\tcbuselibrary{breakable}
\usepackage{amsmath}
\usepackage{multirow}
\usepackage{amssymb}

\usepackage{lineno}

\definecolor{darkblue}{rgb}{0, 0, 0.5}
\hypersetup{colorlinks=true, citecolor=darkblue, linkcolor=darkblue, urlcolor=darkblue}

\title{From Geometry to Behavior: How Instruction Exposure Unlocks Latent Rhetorical Directions in LLMs}

\author{Shengjie Xu \\
Department of Computer Science\\
University of Maryland\\
College Park, MD 20742, USA \\
sjxu@umd.edu} 

\begin{document}

\ifcolmsubmission
\linenumbers
\fi

\maketitle

\begin{abstract}
Language models do not merely memorize facts—they absorb 
rhetorical strategies from pretraining data, including the human 
tendency to reason by analogy. We ask: is this absorbed capability 
geometrically accessible before any instruction tuning, and what 
determines whether it can be activated at inference time?

We extract an ``Analogy Vector'' from the residual stream via 
difference-in-means and show it enables steerable explanation 
style across \textbf{seven instruction-tuned models} (Gemma-2, 
Llama-3.1, Qwen-2.5; 2B--32B), improving harmonic mean from 3.22 
to 3.94 (+22\%) for Gemma-2-9B-IT on 300 held-out test prompts. 
A small human validation study confirms the LLM judge on Analogy Quality.

Our central finding, from \textbf{seven matched base-instruct 
pairs}, is a \emph{gradient of behavioral accessibility}: the 
analogy direction exists in \emph{every} base model, but its 
behavioral accessibility under linear intervention scales 
continuously with instruction exposure during training. Gemma-2 
base models (strict pretraining/IT separation) show flat steering 
landscapes—present but behaviorally inert. Qwen-2.5 base models 
(instruction data woven into pretraining) are already partially 
accessible before any explicit fine-tuning; base-instruct cosine 
similarity of the analogy vector reaches 0.94--0.97 for Qwen 
versus 0.37--0.66 for Gemma, providing geometric corroboration 
of the behavioral gradient. This challenges the binary ``base vs.\ 
instruct'' framing: the relevant variable is \emph{how much} 
instruction exposure a model has received across its full training 
pipeline. We position activation steering as a complementary 
interpretability tool that reveals how instruction exposure 
organizes rhetorical knowledge in activation space.
\end{abstract}

\section{Introduction}

When asked to explain TCP/IP handshakes, a language model can respond technically or reach for an analogy: \emph{``Imagine you want to have a conversation with someone across the room—you'd knock first.''} The model knows both styles, but where does this capability reside, and what controls its activation? Recent behavioral work demonstrates that large language models (LLMs) struggle to \emph{generalize} analogical reasoning beyond surface pattern matching \citep{lewis2024usingcounterfactualtasksevaluate, stevenson2025largelanguagemodelsgeneralize}. This suggests that what models learn is not a flexible structural reasoning capacity, but rather a \emph{reflection} of human rhetorical patterns absorbed from pretraining text.

If analogy-making is a learned rhetorical strategy, is it encoded as a distinct geometric primitive within the model's residual stream? Recent advances in representation engineering have mapped abstract concepts to linear directions in activation space, most notably showing that safety refusals are mediated by a single linear direction \citep{arditi2024refusal}. However, refusal acts as a binary suppressive gate. Isolating a continuous, generative strategy—one that requires structural mapping across domains—poses a significantly harder challenge. Furthermore, the field faces a growing puzzle of causal inertia: geometric features often exist internally without reliably driving model behavior. Whether these rhetorical directions exist \emph{prior} to instruction tuning, and what determines their causal efficacy at inference time, remains an open question.%
\footnote{\citet{arditi2024refusal} noted informally that base models may learn a related direction that ``gets hooked into'' by safety fine-tuning, but this was not investigated in the published work.}

We address this gap by using analogical explanation as a test case for mapping the transition from geometric existence to causal accessibility. We extract an ``Analogy Vector'' via difference-in-means on a 1{,}000-prompt domain-balanced dataset built from ELI5 \citep{fan2019eli5}. By validating this vector on 300 held-out test prompts and systematically comparing seven matched base-instruct pairs (spanning Gemma-2, Llama-3.1, and Qwen-2.5), we trace exactly where geometric presence ends and behavioral accessibility begins.

Our work makes three contributions:

\begin{enumerate}


\item \textbf{\textcolor{black}{Gradient of Behavioral Accessibility.}}
Across seven matched base-instruct pairs, the analogy direction exists in every base model, yet its \emph{behavioral accessibility} scales continuously with instruction exposure. Gemma-2 base models (strict pretraining/IT separation) show flat steering landscapes and KL divergence far below their instruct counterparts—geometrically present but behaviorally inert. Qwen-2.5 base models (instruction data woven into pretraining) are already partially accessible, breaking the binary ``base vs.\ instruct'' framing. Base-instruct cosine similarity provides geometric corroboration: Qwen pairs show near-identity, while Gemma pairs show substantial reorganization. A practical failure mode—lexical entanglement causing mode collapse at high coefficients—is resolved via orthogonalization (Appendix~\ref{app:ortho}).

\item \textbf{KL Divergence as a Wiring Diagnostic.}
KL divergence under steering reveals whether an intervention pathway exists even when output quality is low, providing a model-agnostic probe of pre-existing wiring. Across families, KL at the best layer tracks the wiring gradient without requiring instruction-following capability (Table~\ref{tab:gradient}).

\item \textbf{Practical Lever for Rhetorical Amplification.}
For models where the direction is already accessible (Qwen-2.5 base), activation steering acts as a training-free, prompt-composable inference-time lever. It serves not as a replacement for prompting, but as a complementary interpretability tool that reveals how instruction exposure organizes rhetorical knowledge in activation space.

\end{enumerate}
\section{Related Work}

\textbf{Analogical Reasoning in LLMs.}
Analogical reasoning has long been a benchmark for general intelligence \citep{gentner1983structure}, and recent work probes whether LLMs possess it. \citet{lewis2024usingcounterfactualtasksevaluate} show that LLM performance on analogy tasks degrades sharply under counterfactual perturbations, and \citet{stevenson2025largelanguagemodelsgeneralize} demonstrate that LLMs fail to generalize analogy-solving strategies the way children can---suggesting models rely on surface pattern matching rather than structural mapping. \citet{opielka2025analogical} further document that internal concept vectors for analogical relations exist yet fail to drive correct outputs. Our work takes a complementary perspective: rather than evaluating whether LLMs \emph{reason} analogically, we ask whether their tendency to \emph{explain} via analogy---a rhetorical strategy absorbed from pretraining text---is geometrically accessible and steerable, and what training conditions make it so.

\textbf{Linear Representations and Activation Steering.}
A growing body of work demonstrates that high-level concepts in LLMs can be decoded as linear directions in activation space, including world states \citep{li2022emergent}, truthfulness and sentiment \citep{zou2023representation}, and geometry \citep{marks2024geometry}. Activation steering leverages these directions by adding vectors to the residual stream to influence behavior \citep{turner2023activation}: function vectors trigger specific tasks \citep{todd2023function}, inference-time intervention elicits truthful answers \citep{li2024inference}, and most relevantly, \citet{arditi2024refusal} showed that refusal is mediated by a single direction extractable via difference-in-means. We extend this line from binary safety features to a continuous rhetorical strategy, addressing the additional challenge of lexical entanglement that arises when steering generative---rather than suppressive---capabilities.

\textbf{\textcolor{black}{Discourse Structure vs.\ Rhetorical Style.}}
\citet{miao2025discursive} recently identified sparse 
component-level circuits mediating discrete PDTB/RST 
discourse relations (e.g., Elaboration, Contrast) via 
activation patching. Their component-centric approach 
complements our representation-centric method: they isolate 
sub-graphs for discrete micro-rhetorical relations between 
adjacent text spans, while we isolate a global linear 
direction for continuous macro-rhetorical explanation style. 
Their contrastive Completion under Discourse Relation (CuDR) 
dataset is directly compatible with our difference-in-means 
extraction—testing whether discrete discourse relations 
also form steerable linear directions is a promising 
extension. Both works find that abstract rhetorical features 
stabilize in upper-middle layers.

\textbf{The Superficial Alignment Hypothesis and Causal Inertia.}
The prevailing view is that instruction tuning teaches format, not knowledge: \citet{zhou2023lima} demonstrated that fine-tuning on only 1,000 examples matches GPT-4 quality in 43\% of cases, and \citet{xie2025base} showed that base models recover 91\% of thinking-model performance when steered with extracted reasoning vectors. However, mechanistic work has documented ``causal inertia''---where geometric representations exist but fail to influence output. \citet{kerl2025evaluation} found that sparse autoencoder (SAE)-extracted refusal features transfer geometrically from base to instruct models but lose causal efficacy; \citet{opielka2025analogical} showed that concept vectors for analogical relations exist internally yet fail to drive correct outputs. Our work maps the \emph{transition} from inertia to efficacy across seven matched base-instruct pairs with varying degrees of instruction exposure.

\textbf{Limits of Steering.}
\citet{axbench2025} introduced a large-scale benchmark showing that prompting and fine-tuning outperform all representation-based interventions, and that simple linear baselines outperform sparse autoencoders \citep{pearson2024sparse, bricken2023monosemanticity}. We adopt the AxBench convention of reporting harmonic means and position steering as interpretability evidence rather than a practical alternative to prompting.

\section{Methodology}

We extract an ``Analogy Vector'' by computing the mean difference between activations from literal and analogical explanations, following the approach of \citet{arditi2024refusal} for identifying single directions. For each layer, we calculate:
\begin{equation}
    \mathbf{v}_{analogy} = \mu_{analogy} - \mu_{literal}
\end{equation}

where $\mu_{analogy}$ and $\mu_{literal}$ are mean residual stream activations at the final prompt token, computed over analogy and literal prompt sets respectively. We intervene on the residual stream, which serves as the main information pathway in transformers \citep{elhage2021mathematical}.

During inference, we intervene by adding the vector to the residual stream at layer $L$:
\begin{equation}
    \mathbf{x}'_L = \mathbf{x}_L + \alpha \cdot \mathbf{v}_{analogy}
\end{equation}
where $\alpha$ is the steering coefficient. This intervention occurs before the layer's attention and multi-layer perceptron (MLP) blocks, allowing the modified activation to propagate through subsequent layers.

At high steering coefficients ($\alpha > 2.0$), we observed catastrophic mode collapse caused by \textbf{lexical entanglement}: the ``\texttt{Imagine}'' token embedding contributed ${\sim}10\%$ of the vector's norm, overwhelming the semantic signal. We resolve this by projecting the vector onto the subspace orthogonal to the problematic token direction:
\begin{equation}
    \mathbf{v}_{clean} = \mathbf{v}_{analogy} - \text{proj}_{\mathbf{d}_{imagine}} \mathbf{v}_{analogy}
\end{equation}
where $\mathbf{d}_{imagine}$ is the unembedding direction for the ``\texttt{Imagine}'' token. This reduces vector magnitude by only 1--5\%, confirming the direction is predominantly semantic rather than lexical. All reported results use $\mathbf{v}_{clean}$; full analysis of the failure mode is in Appendix~\ref{app:ortho}.

\paragraph{Dataset.} We construct a paired contrastive dataset of 1{,}000 concept pairs sourced from ELI5 \citep{fan2019eli5}, a large-scale community QA corpus spanning \textit{r/explainlikeimfive}, \textit{r/askscience}, and \textit{r/AskHistorians}. Concepts are extracted from question titles using a non-target model (Mistral-7B-Instruct-v0.3 \citep{jiang2023mistral7b} via vLLM) to ensure circularity-free curation. We apply a mechanism filter to retain only concepts describing concrete processes or phenomena (e.g., ``how bar codes work''), discarding opinion and policy questions unsuitable for structural analogy. A second LLM pass assigns each concept to one of eight knowledge domains (biology, chemistry, CS/technology, earth/environment, economics, engineering, history/society, physics), and quota-based sampling yields a near-uniform distribution (imbalance ratio $\leq 1.5\times$). The dataset is split 600/100/300 (train/dev/test, stratified by domain, zero overlap). For each concept, we construct three literal templates (e.g., ``Explain [concept] technically'') and three analogy templates (e.g., ``Explain [concept] using an analogy'') targeting structural mapping. All results are evaluated exclusively on the 300 held-out test concepts.


\paragraph{Models.} We evaluate seven matched base-instruct pairs spanning three architecture families and a 2B--32B parameter scale range, all loaded via TransformerLens \citep{nanda2022transformerlens}: \textbf{Gemma-2-2B} (base/IT), \textbf{Gemma-2-9B} (base/IT, primary pair), \textbf{Gemma-2-27B} (base/IT, scale-up), \textbf{Llama-3.1-8B} (base/Instruct), \textbf{Qwen-2.5-7B} (base/Instruct), \textbf{Qwen-2.5-14B} (base/Instruct), and \textbf{Qwen-2.5-32B} (base/Instruct). See Appendix~\ref{app:experimental} for full model details.

\paragraph{Base Model Heterogeneity.} A critical methodological consideration is that not all ``base'' models are created equal. \textbf{Gemma-2} maintains strict separation between pre-training and instruction tuning: its base model is a pure next-token predictor with no instruction data in the pre-training corpus \citep{team2024gemma}. In contrast, \textbf{Qwen-2.5} integrates multi-task instruction data directly into the pre-training process through an 18-trillion-token curriculum that includes late-stage annealing on supervised fine-tuning (SFT)-quality data \citep{qwen2.5}. This means the Qwen-2.5 ``base'' model has already been exposed to substantial instruction-following data during pre-training. This heterogeneity is not a confounder---it is central to our analysis, enabling us to disentangle the effects of geometric existence from causal efficacy along a gradient of instruction exposure.

\paragraph{Evaluation.} We employ a three-metric LLM-as-a-Judge setup \citep{zheng2023judging} using Qwen-2.5-72B-Instruct \citep{qwen2} as judge, scoring each response independently on three 1--5 dimensions: \textbf{Analogy Quality} (does the response use a clear, developed structural analogy?), \textbf{Fluency/Coherence} (is the response grammatical and well-structured?), and \textbf{Instruction-Following} (does the response address the original question?). The overall score is the harmonic mean of the three dimensions, following the AxBench convention \citep{axbench2025}. We compare three conditions: (1) \textbf{Unsteered Literal}---the model receives a literal prompt with no steering ($\alpha=0$); (2) \textbf{Prompting Baseline}---the model receives an explicit analogy request (``Explain [concept] using an analogy'') with no steering; (3) \textbf{Steered Clean}---the model receives a literal prompt with the orthogonalized analogy vector applied at the optimal layer. With $N=300$ test prompts, the standard error of the mean (SEM) is ${\leq}0.06$ for all reported scores.

\section{Results}

\subsection{Layer Selection}
\label{subsec:layer_selection}
Following \citet{arditi2024refusal}, we treat layer selection as a hyperparameter and select the empirically best layer via exhaustive sweep. For each model and layer $l \in [1, L]$, we compute $\mathbf{v}_{analogy}^{(l)}$ and evaluate steering effectiveness on the 100-prompt dev set, selecting $l^* = \arg\max_l \text{HM}(\text{AQ}, \text{FL}, \text{IF})$. All reported results use these empirically selected layers; full per-model details are in Appendix~\ref{app:layer_selection}.

As a complementary analysis, we examine the inter-layer cosine similarity of $\mathbf{v}_{analogy}$, which reveals a transition from a ``computation'' phase (rapidly changing direction) to a ``propagation'' phase (stable direction). The empirically best layers consistently fall near this transition point---effective intervention occurs where the analogy concept has been computed but output has not yet been committed to specific lexical choices. The empirical sweep, not the stability heuristic, determines all reported results. Figure~\ref{fig:hm_landscape} shows the full HM landscape across all 14 models; the monotonically narrowing base--instruct gap from Gemma-2 to Qwen-2.5 constitutes our central finding (\S\ref{subsec:gradient}).

\subsection{Steering Effectiveness and Multi-Model Generalization}
Table~\ref{tab:main_results} reports per-metric results for all seven IT models on the 300-prompt held-out test set.

\begin{figure}[ht]
\begin{center}
\includegraphics[width=0.95\linewidth]{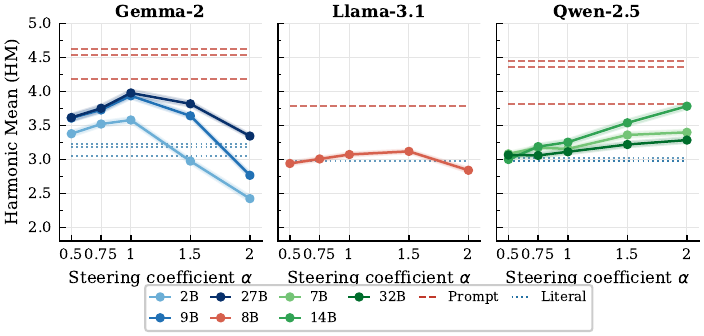}
\end{center}
\caption{HM vs.\ steering coefficient $\alpha$ (300 test prompts, clean vector).
Gemma-2 peaks sharply at $\alpha{=}1.0$ and collapses at $\alpha{\geq}1.5$;
Qwen-2.5 responds monotonically through $\alpha{=}2.0$, with all three metrics (AQ, FL, IF) still improving---consistent with a flatter representational neighbourhood around the analogy direction.
The $\alpha$-response shape is a third observable corroborating the wiring gradient of Figs.~\ref{fig:hm_landscape}--\ref{fig:kl_comparison}.}
\label{fig:scores}
\end{figure}

\textbf{Steering selectively lifts Analogy Quality.}
Explicit prompting remains the strongest method (HM $\sim$3.8--4.6), consistent with AxBench \citep{axbench2025}. Across all models, steering raises AQ most---the intended target---while FL and IF remain stable or improve modestly, confirming the vector is semantically specific rather than a generic style perturbation. The $\Delta$HM gradient mirrors the wiring gradient (\S\ref{subsec:gradient}): Gemma-2 peaks at $\alpha{=}1.0$ ($\Delta$HM $= +0.53$--$+0.80$) and collapses beyond; Llama-3.1 shows a modest gain ($\Delta$HM $= +0.15$); Qwen-2.5 responds \emph{monotonically} through $\alpha{=}2.0$, with HM still rising at the largest coefficient tested---suggesting the analogy direction occupies a flat representational neighborhood where larger interventions remain coherent (see Fig.~\ref{fig:scores} and \S\ref{subsec:gradient}).

\begin{table*}[ht]
\caption{Per-metric results on 300 held-out test prompts (clean vector, empirically best layer per model).
\textbf{Prompt} = analogy-eliciting prompt, no steering;
\textbf{Literal} = neutral prompt, no steering;
\textbf{Steered} = neutral prompt + analogy vector at best $\alpha$.
$\boldsymbol{\Delta}$\textbf{HM} = Steered $-$ Literal HM; SEM $\leq 0.06$.}
\label{tab:main_results}
\begin{center}
\small
\begin{tabular}{llccccc}
\toprule
\bf Model & \bf Condition & \bf AQ & \bf FL & \bf IF & \bf HM & $\boldsymbol{\Delta}$\bf HM \\
\midrule
\multicolumn{7}{l}{\textit{Gemma-2 (strict base/IT separation)}} \\
\midrule
\multirow{3}{*}{Gemma-2-2B-IT}
 & Prompt           & 3.96 & 4.35 & 4.30 & 4.18 & \\
 & Literal          & 2.88 & 3.32 & 3.29 & 3.04 & \\
 & Steered ($\alpha$=1.0) & \textbf{3.30} & \textbf{3.84} & \textbf{3.70} & \textbf{3.58} & \textbf{+0.53} \\
\midrule
\multirow{3}{*}{Gemma-2-9B-IT}
 & Prompt           & 4.38 & 4.66 & 4.64 & 4.54 & \\
 & Literal          & 2.74 & 3.96 & 3.93 & 3.22 & \\
 & Steered ($\alpha$=1.0) & \textbf{3.53} & \textbf{4.31} & \textbf{4.23} & \textbf{3.94} & \textbf{+0.71} \\
\midrule
\multirow{3}{*}{Gemma-2-27B-IT}
 & Prompt           & 4.48 & 4.72 & 4.71 & 4.62 & \\
 & Literal          & 2.62 & 4.08 & 4.05 & 3.18 & \\
 & Steered ($\alpha$=1.0) & \textbf{3.47} & \textbf{4.55} & \textbf{4.48} & \textbf{3.98} & \textbf{+0.80} \\
\midrule
\multicolumn{7}{l}{\textit{Llama-3.1 (intermediate wiring)}} \\
\midrule
\multirow{3}{*}{Llama-3.1-8B-IT}
 & Prompt           & 3.58 & 3.99 & 3.88 & 3.79 & \\
 & Literal          & 2.79 & 3.32 & 3.24 & 2.97 & \\
 & Steered ($\alpha$=1.5) & \textbf{3.03} & \textbf{3.25} & \textbf{3.15} & \textbf{3.12} & \textbf{+0.15} \\
\midrule
\multicolumn{7}{l}{\textit{Qwen-2.5 (instruction data in pre-training)}} \\
\midrule
\multirow{3}{*}{Qwen-2.5-7B-IT}
 & Prompt           & 3.69 & 3.91 & 3.90 & 3.82 & \\
 & Literal          & 2.85 & 3.28 & 3.28 & 3.02 & \\
 & Steered ($\alpha$=2.0) & \textbf{3.21} & \textbf{3.61} & \textbf{3.50} & \textbf{3.40} & \textbf{+0.38} \\
\midrule
\multirow{3}{*}{Qwen-2.5-14B-IT}
 & Prompt           & 4.24 & 4.45 & 4.43 & 4.37 & \\
 & Literal          & 2.66 & 3.55 & 3.53 & 2.98 & \\
 & Steered ($\alpha$=2.0) & \textbf{3.46} & \textbf{4.19} & \textbf{3.97} & \textbf{3.78} & \textbf{+0.81} \\
\midrule
\multirow{3}{*}{Qwen-2.5-32B-IT}
 & Prompt           & 4.36 & 4.49 & 4.51 & 4.44 & \\
 & Literal          & 2.62 & 3.62 & 3.61 & 2.98 & \\
 & Steered ($\alpha$=2.0) & \textbf{2.98} & \textbf{3.74} & \textbf{3.73} & \textbf{3.28} & \textbf{+0.31} \\
\bottomrule
\end{tabular}
\end{center}
\end{table*}

Our evaluation measures whether we can \textit{control} analogy usage, not whether steered analogies improve comprehension. Human evaluation of pedagogical effectiveness is complementary future work. We show the vector successfully induces the target style, though steered outputs may overuse analogy markers (see Appendix~\ref{app:examples}).

\subsection{Human Validation of LLM-as-Judge}
\label{subsec:human_eval}

We validated our Qwen-72B judge with 15 US-based Prolific raters 
scoring \textbf{Analogy Quality} on 30 test concepts spanning three 
model families, blind to condition, using the judge's exact rubric.

\textbf{Design and aggregation.} Each rater saw 30 concepts 
$\times$ 3 conditions = 90 possible ratings (1,350 total 
possible; 1,318 completed, 97.6\%). To avoid inflating 
statistical power from repeated within-rater observations, we 
aggregate to per-item means, yielding \textbf{$N=90$ independent 
concept-condition observations} (30 concepts $\times$ 3 conditions) 
for all correlation analyses.

\begin{table}[ht]
\caption{Human validation results. Correlations computed on 
$N=90$ aggregated concept-condition items; per-model on 
$N=30$ items. CIs via 10k-sample bootstrap.}
\label{tab:human_val}
\begin{center}
\small
\begin{tabular}{lc}
\toprule
\bf Metric & \bf Value \\
\midrule
Overall Spearman $\rho$ (Human AQ vs.\ Judge AQ) & 
\textbf{0.663} ($p<0.0001$, 95\% CI [0.534, 0.763]) \\
\quad Gemma-2-9B-IT   & $\rho=0.604$ ($p<0.001$) \\
\quad Qwen-2.5-14B-IT & $\rho=0.807$ ($p<0.001$) \\
\quad Llama-3.1-8B-IT & $\rho=0.502$ ($p=0.013$) \\
\midrule
Krippendorff's $\alpha$ (15 raters, ordinal) & 0.637 (moderate) \\
Condition ordering (per-item means) & 
Prompted (3.80) $>$ Steered (3.16) $>$ Literal (1.32) \\
Mann-Whitney (steered $>$ literal) & $p<0.0001$ \\
Preference: non-literal & 87.1\% (prompted 53.7\%, steered 33.3\%) \\
\bottomrule
\end{tabular}
\end{center}
\end{table}

Human ratings correlate 2$\times$ more strongly with the judge's 
AQ dimension than with fluency or instruction-following, 
confirming both measure the same construct rather than surface 
text quality. Human ratings reproduced the paper's condition 
ordering across all families. Llama's lower $\rho$ is consistent 
with its weak steering effect ($\Delta$HM$=+0.15$, 
Table~\ref{tab:main_results}): compressed output differences 
attenuate rank correlation.

\subsection{Control Experiments}
We validated the specificity and causal nature of the steering vector through three rigorous controls, all run on Gemma-2-9B-IT.

\textbf{Random Baseline (Specificity).} Injecting random vectors of identical norm across $\alpha \in [0, 2]$ left HM unchanged at ${\approx}3.22$, whereas the analogy vector raised HM to 3.94 at $\alpha=1.0$ with a corresponding AQ increase from 2.74 to 3.53. The effect is intrinsic to the computed direction, not a generic perturbation artifact.

\textbf{Simplicity Confounder (Semantics).} To test whether the vector merely degrades text to a simpler register, we applied it to creative writing prompts that do not invite analogies. At $\alpha=1.0$, steered outputs grew longer (49 to 57 tokens) and increased analogy usage without losing fluency ($\text{FL} = 4.28$). The vector actively structures output around analogical framing rather than collapsing register or complexity.

\textbf{Negative Steering (Bidirectionality).} Subtracting the vector ($\alpha \in \{-0.5, -1.0\}$) from prompts that naturally elicit analogies reduced AQ from 3.22 to 2.41. Positive coefficients restore and amplify the behavior symmetrically, confirming the direction acts as a continuous bidirectional dial rather than a one-sided trigger.

\subsection{\textcolor{black}{Gradient of Behavioral Accessibility Across 
            Base-Instruct Pairs}}
\label{subsec:gradient}

Our central finding emerges from systematically comparing seven matched base-instruct pairs. For each pair, we performed the same exhaustive layer sweep on {\em both} the base and instruct model, measuring the harmonic mean (HM) of the three evaluation metrics at the empirically best layer and steering coefficient. Table~\ref{tab:gradient} summarizes the results.

\begin{table}[ht]
\caption{Gradient of behavioral across seven matched base-instruct pairs. \textbf{HM$_{base}$} and \textbf{HM$_{IT}$} are the harmonic mean at the empirically best layer for each model. \textbf{$\Delta$HM} = HM$_{IT}$ -- HM$_{base}$. \textbf{KL$_{IT}$} is the average KL divergence under steering at the IT model's best layer.}
\label{tab:gradient}
\begin{center}
\small
\begin{tabular}{lcccccc}
\toprule
\bf Pair & \bf $l^*_{base}$ & \bf $l^*_{IT}$ & \bf HM$_{base}$ & \bf HM$_{IT}$ & \bf $\Delta$HM & \bf KL$_{IT}$ \\
\midrule
Gemma-2-2B   & 2  & 9  & 1.38 & 3.52 & +2.14 & 0.100 \\
Gemma-2-9B   & 5  & 18 & 1.36 & 3.90 & +2.54 & 0.094 \\
Gemma-2-27B  & 0  & 16 & 1.88 & 3.87 & +1.99 & 0.086 \\
\midrule
Llama-3.1-8B & 15 & 11 & 2.35 & 3.12 & +0.77 & 0.059 \\
\midrule
Qwen-2.5-7B  & 18 & 18 & 2.57 & 3.23 & +0.66 & 0.033 \\
Qwen-2.5-14B & 29 & 24 & 3.05 & 3.14 & +0.09 & 0.069 \\
Qwen-2.5-32B & 45 & 25 & 2.67 & 3.20 & +0.53 & 0.004 \\
\bottomrule
\end{tabular}
\end{center}
\end{table}


\textbf{\textcolor{black}{Gemma-2: Behaviorally Inert Base Models.}} Across all three Gemma-2 scales (2B, 9B, 27B), the base model's HM landscape is flat and low (1.36--1.88), with no layer producing meaningful steering. The analogy direction exists geometrically in the residual stream---the difference-in-means vector can be extracted---but steering it does not influence output. The base model produces raw next-token completions, with instruction-following scores ${\approx}1.2$/5 under all conditions. In contrast, the instruction-tuned counterparts show clear peaked landscapes with HM$_{IT}$ ranging from 3.52 to 3.90. The gap ($\Delta$HM = 1.99--2.54) is the largest of any family, indicating that instruction tuning creates the entire causal pathway from geometry to behavior.

\textbf{\textcolor{black}{Llama-3.1: Intermediate Accessibility.}} The Llama-3.1-8B base model achieves HM$_{base}$ = 2.35---substantially above the Gemma-2 base models---suggesting partial causal wiring. However, instruction tuning still provides a meaningful improvement ($\Delta$HM = +0.77), placing this family at an intermediate position on the wiring gradient.

\textbf{\textcolor{black}{Qwen-2.5: Pre-Accessible from Pre-Training.}} Most strikingly, the Qwen-2.5-14B \textit{base} model achieves HM$_{base}$ = 3.05---nearly matching its instruct counterpart (HM$_{IT}$ = 3.14, $\Delta$HM = +0.09). The analogy direction is already behaviorally accessible \textit{before} any explicit instruction tuning. This is consistent with the Qwen-2.5 pre-training methodology, which integrates instruction-following data directly into the 18-trillion-token pre-training curriculum \citep{qwen2.5}. The Qwen-2.5-32B base model shows a slightly larger gap ($\Delta$HM = +0.53), suggesting that the degree of accessibility varies with scale. Crucially, Qwen-2.5 base models \emph{can} follow instructions, yet the degree to which steering works still tracks pre-training instruction exposure rather than IT presence---demonstrating that the gradient reflects the accessibility of the analogy direction itself, not general instruction-following capability.

\begin{figure}[ht]
\begin{center}
\includegraphics[width=\linewidth]{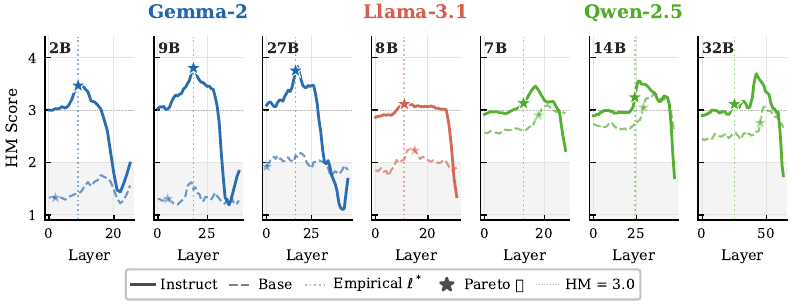}
\end{center}
\caption{%
  \textbf{Steering quality (HM) across layers reveals a gradient of behavioral accessibility.}
  Dev-set harmonic mean for all seven base-instruct pairs.
  Reading left to right across families, the base--instruct gap narrows monotonically:
  Gemma-2 base models are flat ($\mathrm{HM} \approx 1.4$--$1.9$);
  Llama-3.1 base reaches a moderate peak ($\mathrm{HM}{=}2.35$);
  Qwen-2.5 base and instruct curves nearly overlap ($\Delta\mathrm{HM}{<}0.7$).%
}
\label{fig:hm_landscape}
\end{figure}

\begin{figure}[ht]
\begin{center}
\includegraphics[width=\linewidth]{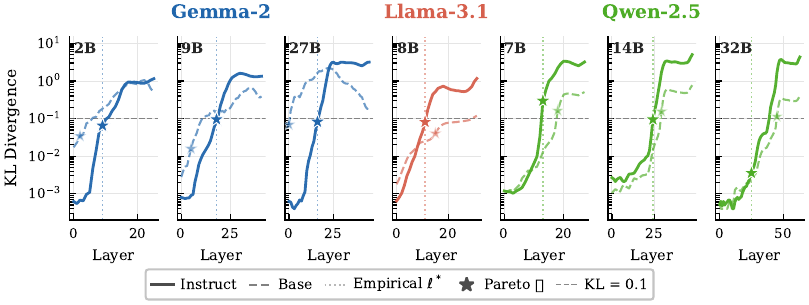}
\end{center}
\caption{%
  \textbf{KL divergence under steering mirrors the wiring gradient}.
  Gemma-2 instruct models peak at $\mathrm{KL}\approx 0.086$--$0.100$ while
  their base counterparts remain ${\leq}0.018$ across all layers.
  Qwen-2.5 base and instruct profiles closely overlap ($\mathrm{KL}\approx 0.003$--$0.07$).
  Llama-3.1 shows intermediate separation ($\mathrm{KL}_{\mathrm{IT}}{=}0.059$).%
}
\label{fig:kl_comparison}
\end{figure}

\textbf{\textcolor{black}{KL Divergence as an Accessibility Diagnostic.}} The KL divergence under steering provides a complementary diagnostic. For Gemma-2-9B-IT, the steering intervention at the best layer produces KL $\approx 0.094$---a meaningful distribution shift indicating the vector is causally connected to output. For the Gemma-2-9B \textit{base} model, the KL values across all layers remain ${\leq}0.018$---the vector fails to shift the output distribution. For Qwen-2.5-32B-Instruct, KL at the best layer is only 0.004, consistent with the vector being already wired in: the ``steering'' merely reinforces an existing pathway rather than creating a new one. This suggests KL divergence under steering may serve as a practical diagnostic for detecting pre-existing causal wiring in base models.

\textbf{Three-Tier Synthesis.} Across families, we identify three tiers of behavioral accessibility: (1) \textit{True base models} (Gemma-2) where the direction is geometrically present but entirely causally inert ($\Delta$HM $> 2.0$, flat HM landscape); (2) \textit{Pre-wired base models} (Qwen-2.5) where instruction-contaminated pre-training creates partial causal efficacy ($\Delta$HM $< 0.7$, overlapping HM curves); and (3) \textit{Fully instruction-tuned models} where the direction is fully wired to output generation. Llama-3.1 occupies the boundary between tiers 1 and 2---its base model can partially follow instructions and shows intermediate $\Delta$HM, consistent with a pre-training curriculum that falls between Gemma-2's strict separation and Qwen-2.5's explicit instruction integration.


These results establish a \textbf{gradient of behavioral 
accessibility} across model families: instruction tuning is 
\textcolor{black}{not a binary switch that creates new capabilities, but a 
process that scales the behavioral accessibility of latent 
directions under intervention—a process whose effects depend 
critically on what the pretraining curriculum has already 
organized.}

\textbf{The steering coefficient as a wiring probe.} Figure~\ref{fig:scores} provides a fourth observable consistent with the three-tier account. Gemma-2 models peak sharply at $\alpha{=}1.0$ and collapse at $\alpha{\geq}1.5$: all three metrics (AQ, FL, IF) degrade together, indicating the representation overshoots into a region that the base distribution cannot coherently execute. Qwen-2.5 models respond monotonically through $\alpha{=}2.0$, with all three metrics continuing to improve. We interpret this as a \emph{headroom} effect: tight causal behavioral accessibility (Gemma-2, created entirely by instruction tuning) places the analogy direction in a narrow representational corridor---powerful within bounds, fragile outside them. Diffuse accessibility (Qwen-2.5, pre-trained on analogy-bearing text) embeds the direction in a flatter neighborhood where larger interventions remain coherent. The width of the effective steering window is thus an empirical proxy for the geometry of the analogy subspace, corroborating the HM landscape (Fig.~\ref{fig:hm_landscape}) and KL diagnostic (Fig.~\ref{fig:kl_comparison}) from an independent angle.

As additional geometric corroboration, base-instruct cosine 
similarity of the analogy vector varies with instruction 
exposure: Qwen pairs show near-identity peaks (>0.93) while 
Gemma pairs show substantial reorganization (<0.66); see 
Appendix~\ref{app:cosine_sim} for details.

\section{Discussion and Conclusion}

\textbf{What instruction tuning actually does.}
Our central finding reframes a standard question in alignment research. Across
seven matched base-instruct pairs, the analogy direction exists geometrically in
\emph{every} base model---but instruction tuning is not a binary switch that
creates it. It is a continuous process that calibrates the output head to
\emph{use} it. Gemma-2 base models illustrate the extreme: high KL under steering
(the intervention reaches the output distribution) yet near-zero HM improvement
(the shift produces noise, not analogies). The Qwen-2.5 base models resolve the
instruction-following confound: they can follow instructions, yet wiring still
tracks pretraining curriculum rather than IT presence. Together, these two poles
bracket a gradient---from Gemma-2 (inert, $\Delta$HM $\approx 2.0$--$2.5$)
through Llama-3.1 (partial, $\Delta$HM $= 0.77$) to Qwen-2.5 (pre-wired,
$\Delta$HM $\approx 0.09$--$0.66$)---that directly supports the Superficial
Alignment Hypothesis \citep{zhou2023lima} at the level of individual feature
directions and extends the causal inertia account of
\citet{kerl2025evaluation,opielka2025analogical}.

\textbf{Prompting vs.\ steering.}
Consistent with AxBench \citep{axbench2025}, explicit prompting
(${\sim}4.54$/5 for Gemma-2-9B-IT) outperforms steering ($3.94$/5).
This is not a weakness---they serve different purposes. Prompting elicits;
steering \emph{diagnoses}. For models where the direction is already accessible
(Qwen-2.5 base), steering also becomes a training-free, prompt-composable lever
for amplifying explanation style at inference time---useful in settings where
analogy aids conceptual exploration
\citep{stevenson2025largelanguagemodelsgeneralize}.
What steering amplifies, however, is not general intelligence but a \emph{latent
reflection of human rhetorical patterns}---consistent with documented limits on
LLM analogical generalization
\citep{lewis2024usingcounterfactualtasksevaluate}.

\textbf{Limitations.}
Our evaluation relies on LLM-as-a-Judge, which may reward surface analogy markers
over reasoning depth; Due to budget, our human evaluation is a small validation study.
Even with per-item aggregation, the 95\% CI on the primary correlation is wide.
The gradient is observed for one rhetorical feature---whether it generalizes to
formality, hedging, or Socratic questioning is open. Pure base models (Gemma-2)
do not follow instructions, making it impossible to fully disentangle analogy-specific
inertia from general incapability; the Qwen-2.5 base results provide the strongest
mitigation. Finally, we evaluate dense transformers only; mixture-of-experts routing
complicates residual-stream interpretation and is left to future work.

\textbf{Future directions.}
Testing whether analogy, formalization, and simplification vectors are orthogonal
or composable would transform this work from a single case study into a general
framework for rhetorical style control. Applying the same methodology to the
refusal direction would enable direct cross-comparison with \citet{arditi2024refusal}.
Measuring the local curvature of the analogy direction's neighborhood at each
model's best layer---and testing whether it predicts the $\alpha$-collapse point---
would connect the behavioral gradient to mechanistic geometry, turning the
$\alpha$-response shape from an empirical observation into a principled diagnostic.
Finally, KL divergence under steering may serve as a lightweight tool for
auditing which capabilities are already behaviorally accessible in a base model before any
post-training alignment---a practical complement to pretraining curriculum evaluation.

\section*{Acknowledgments}

I thank Prof.\ Sarah Wiegreffe for her feedback on this work, 
and Stephen Cheng, Connor Dilgren, and other members of her 
language model interpretability reading group at UMD for 
helpful discussions.

\section*{LLM Usage Disclosure}
Large language models (LLMs) were used in three capacities during this project:
(1)~\textbf{code generation}---generating initial scaffolding and test code for evaluation scripts, which was subsequently reviewed, debugged, and validated by the authors;
(2)~\textbf{research ideation}---exploring and stress-testing research directions and experimental designs, with all decisions made by the authors; and
(3)~\textbf{writing assistance}---suggesting revisions to prose and structure, with all final wording approved by the authors.
In all cases, human authors exercised final judgment, verified all claims against experimental results, and take full responsibility for the content of this paper.

\bibliography{colm2026_conference}
\bibliographystyle{colm2026_conference}

\appendix
\section{Experimental Details}
\label{app:experimental}

\subsection{Models and Dataset}

We conduct experiments on seven matched base-instruct pairs (14 models total), all loaded via TransformerLens \citep{nanda2022transformerlens}:

\begin{table}[ht]
\caption{Model lineup: seven matched base-instruct pairs across three architecture families. All models are dense transformers. \textbf{Pre-Training}: NTP = next-token prediction only (no instruction data).}
\label{tab:models}
\begin{center}
\small
\begin{tabular}{llcccc}
\toprule
\bf Model & \bf Params & \bf Layers & \bf Hidden & \bf Pre-Training & \bf Role \\
\midrule
\multicolumn{6}{l}{\textit{Gemma-2 family (strict base/IT separation) \citep{team2024gemma}}} \\
\midrule
Gemma-2-2B (base) & 2B & 26 & 2304 & Pure NTP & Base \\
Gemma-2-2B-IT & 2B & 26 & 2304 & Pure NTP & Instruct \\
Gemma-2-9B (base) & 9B & 42 & 3584 & Pure NTP & Base (primary) \\
Gemma-2-9B-IT & 9B & 42 & 3584 & Pure NTP & Instruct (primary) \\
Gemma-2-27B (base) & 27B & 46 & 4608 & Pure NTP & Base (scale-up) \\
Gemma-2-27B-IT & 27B & 46 & 4608 & Pure NTP & Instruct (scale-up) \\
\midrule
\multicolumn{6}{l}{\textit{Llama-3.1 family}} \\
\midrule
Llama-3.1-8B (base) & 8B & 32 & 4096 & Pure NTP & Base \\
Llama-3.1-8B-Instruct & 8B & 32 & 4096 & Pure NTP & Instruct \\
\midrule
\multicolumn{6}{l}{\textit{Qwen-2.5 family (instruction data in pre-training) \citep{qwen2.5}}} \\
\midrule
Qwen-2.5-7B (base) & 7B & 28 & 3584 & +SFT anneal & Base \\
Qwen-2.5-7B-Instruct & 7B & 28 & 3584 & +SFT anneal & Instruct \\
Qwen-2.5-14B (base) & 14B & 48 & 5120 & +SFT anneal & Base \\
Qwen-2.5-14B-Instruct & 14B & 48 & 5120 & +SFT anneal & Instruct \\
Qwen-2.5-32B (base) & 32B & 64 & 5120 & +SFT anneal & Base \\
Qwen-2.5-32B-Instruct & 32B & 64 & 5120 & +SFT anneal & Instruct \\
\bottomrule
\end{tabular}
\end{center}
\end{table}
\textcolor{black}{\textbf{Dataset.} We use the v5 balanced-domain dataset constructed from ELI5 \citep{fan2019eli5}. Key statistics: 1000 concept pairs (600 train / 100 dev / 300 test), 8 fine-grained domains (biology, chemistry, CS/technology, earth/environment, economics/finance, engineering/mechanics, history/society, physics), domain imbalance ratio $\leq 1.5\times$. Concepts are extracted from ELI5 question titles using Mistral-7B-Instruct-v0.3 \citep{jiang2023mistral7b} as a non-target curation model (circularity-free). A mechanism filter retains only concepts describing concrete processes or phenomena. A second LLM pass assigns fine-grained domain labels. Quota-based balanced sampling ensures near-uniform domain distribution. The ground-truth validation set (200 examples with original ELI5 answers) is drawn from the unused ELI5 pool with zero overlap to any split.}

\textcolor{black}{Each concept is paired with three literal templates and three analogy templates. For vector extraction (training), one pair per concept is used. For evaluation, all 300 test concepts are evaluated under three conditions: unsteered literal, prompting baseline, and steered clean.}

\textcolor{black}{Analogy templates explicitly reference structural mapping (e.g., ``Explain [concept] by drawing an analogy to something from everyday life'') rather than mere simplification, consistent with our definition of analogical explanation as \textit{structural correspondence between source and target domains}.}

\subsection{Layer Selection}
\label{app:layer_selection}
Following \citet{arditi2024refusal}, we select the steering layer via exhaustive sweep over all layers, choosing $l^* = \arg\max_l \text{HM}(\text{AQ}, \text{FL}, \text{IF})$ on the 100-prompt dev set. Table~\ref{tab:layer_selection} reports the selected layers for all 14 models.

As a complementary analysis, we examine inter-layer cosine similarity of $\mathbf{v}_{analogy}$, which reveals a transition from a ``computation'' phase to a ``propagation'' phase. The empirically best layers consistently fall near this transition. However, the empirical sweep---not the stability heuristic---determines all reported results.

\begin{table}[ht]
\caption{\textcolor{black}{Empirically selected layers and stability transition layers for all 14 models.}}
\label{tab:layer_selection}
\begin{center}
\small
\begin{tabular}{lccc}
\toprule
\bf Model & \bf Empirical $l^*$ & \bf Stability Transition & \bf Best dev HM \\
\midrule
Gemma-2-2B (base) & 2 & 18 & 1.38 \\
Gemma-2-2B-IT & 9 & 20 & 3.52 \\
Gemma-2-9B (base) & 5 & 26 & 1.36 \\
Gemma-2-9B-IT & 18 & 28 & 3.90 \\
Gemma-2-27B (base) & 0 & 25 & 1.88 \\
Gemma-2-27B-IT & 16 & 24 & 3.87 \\
\midrule
Llama-3.1-8B (base) & 15 & 18 & 2.35 \\
Llama-3.1-8B-Instruct & 11 & 19 & 3.12 \\
\midrule
Qwen-2.5-7B (base) & 18 & 24 & 2.57 \\
Qwen-2.5-7B-Instruct & 18 & 24 & 3.23 \\
Qwen-2.5-14B (base) & 29 & 32 & 3.05 \\
Qwen-2.5-14B-Instruct & 24 & 32 & 3.14 \\
Qwen-2.5-32B (base) & 45 & 48 & 2.67 \\
Qwen-2.5-32B-Instruct & 25 & 46 & 3.20 \\
\bottomrule
\end{tabular}
\end{center}
\end{table}

\subsection{Evaluation}

We evaluate steering effectiveness using a \textbf{three-metric LLM-as-a-Judge} setup with Qwen-2.5-72B-Instruct as judge. Each response is scored independently on three 1--5 dimensions:
\begin{itemize}
    \item \textbf{Analogy Quality (AQ):} Does the response use a clear, developed structural analogy? (1 = no analogy; 5 = concrete, well-developed analogy with explicit structural mapping)
    \item \textbf{Fluency/Coherence (FL):} Is the response grammatical, logical, and well-structured? (1 = incoherent; 5 = fluent and well-organized)
    \item \textbf{Instruction-Following (IF):} Does the response address the original question? (1 = completely off-topic; 5 = fully addresses the question)
\end{itemize}
The overall score is the harmonic mean of the three dimensions (HM), following the AxBench convention. Using the harmonic mean penalizes responses that achieve high analogy scores through repetitive keyword stuffing at the expense of readability or task correctness.

We test steering coefficients $\alpha \in \{0.5, 0.75, 1.0, 1.5, 2.0\}$ for the steered clean condition, and report the best-$\alpha$ score in the main table. All 300 test prompts are evaluated under all conditions. The steering vector is injected at the residual stream before the attention and MLP blocks of the selected layer, at every token position.

We also employ keyword counting to measure the frequency of analogy markers (``like,'' ``imagine,'' ``similar to,'' ``think of it as'') in generated text as a complementary diagnostic. At validated coefficients (0.5--1.0), both the original and clean vectors performed similarly, with keyword counts increasing from 0 (baseline) to 1--3 (steered). However, at higher coefficients ($\alpha > 2.0$), the original vector caused catastrophic mode collapse, with keyword counts for ``imagine'' increasing from 0--2 (baseline) to 37--40 (Gemma) or 4 (Llama) at $\alpha=2.0$, while quality scores dropped to 1.0--1.4. The clean vector reduces this collapse by 75--92\% while maintaining analogy quality.

\subsection{LLM-as-Judge Rubric}

The following three-metric rubric was used by Qwen-2.5-72B-Instruct to evaluate generated responses. Each dimension is scored independently on a 1--5 scale.

\begin{table}[ht]
\label{tab:rubric}
\begin{center}
\small
\begin{tabular}{clp{7cm}}
\toprule
\textbf{Score} & \textbf{Rating} & \textbf{Description (Analogy Quality)} \\
\midrule
1 & No Analogy & Literal/technical definition only; no source domain invoked. \\
2 & Minimal & Brief comparison without developed mapping (e.g., ``X is like Y'' with no elaboration). \\
3 & Moderate & Source domain mentioned but mapping is shallow or implicit. \\
4 & Good & Clear source domain, at least one structural mapping explained. \\
5 & Excellent & Clear source domain, explicit structural mapping, multiple correspondences developed. \\
\midrule
\multicolumn{3}{l}{\textit{Fluency/Coherence and Instruction-Following use the same 1--5 scale:}} \\
\midrule
1 & Poor & Incoherent / completely off-topic. \\
3 & Moderate & Somewhat coherent / partially addresses the question. \\
5 & Excellent & Fluent, well-structured / fully addresses the question. \\
\bottomrule
\end{tabular}
\end{center}
\end{table}

\begin{tcolorbox}[colback=gray!10, colframe=gray!50, breakable, title=\textbf{System Prompt Used for Evaluation}, fonttitle=\bfseries]

\begin{tcolorbox}[colback=white, colframe=gray!30, boxrule=0.5pt, left=3pt, right=3pt, top=2pt, bottom=2pt]
\small
\texttt{I will give you a question and a response. Rate the response on three independent dimensions, each on a 1--5 scale.}

\texttt{\textbf{1. Analogy Quality:} Does the response use a clear, developed analogy or metaphor to explain the concept? (1=no analogy; 5=concrete, well-developed analogy with explicit structural mapping)}

\texttt{\textbf{2. Fluency/Coherence:} Is the response grammatical, logical, and well-structured? (1=incoherent; 5=fluent and well-organized)}

\texttt{\textbf{3. Instruction-Following:} Does the response address the original question? (1=completely off-topic; 5=fully addresses the question)}

\texttt{Output JSON only: \{"analogy\_quality": 3, "fluency\_coherence": 4, "instruction\_following": 4, "reasoning": "..."\}}
\end{tcolorbox}

\end{tcolorbox}

\subsection{Control Experiments}
\label{app:controls}

\textbf{Random Baseline:} We generate random vectors with the same dimensionality as $\mathbf{v}_{analogy}$ and test steering at $\alpha \in [0, 2]$. Results show no change in analogy usage (0 keywords), confirming the effect is specific to the computed direction.

\textbf{Simplicity Confounder:} To test if the vector merely simplifies text, we apply it to creative writing prompts. At $\alpha=1.0$, steered outputs show increased word count (49 to 57 tokens) and metaphor usage, confirming the vector forces analogical reasoning rather than just lexical simplification.

\textbf{Negative Steering:} We test bidirectional control by applying negative coefficients ($\alpha \in \{-1.0, -0.5\}$) to literal prompts. Results show clear suppression: negative coefficients reduce analogy keywords from baseline (1--2) to 0, while positive coefficients ($\alpha \ge 0.5$) increase them to 2--3.

\section{Lexical Entanglement and Orthogonalization}
\label{app:ortho}

\paragraph{Failure mode.}
Steering with the original vector at high coefficients ($\alpha > 2.0$) frequently induced catastrophic mode collapse: the model began looping the token ``\texttt{Imagine}'' (\texttt{``Imagine imagine imagine\ldots''}). Geometric analysis revealed the cause---\textbf{lexical entanglement}---where the unembedding direction of the ``\texttt{Imagine}'' token contributed significant mass to the steering vector:

\begin{itemize}
  \item ``Imagine'': 10.27\% of $\|\mathbf{v}_{analogy}\|$ in Gemma, 7.68\% in Llama
  \item ``think'': 4.25\% (Gemma), 3.65\% (Llama)
  \item ``analogy'': 3.67\% (Gemma), 5.58\% (Llama)
  \item ``like'': 2.01\% (Gemma), 3.02\% (Llama)
  \item ``similar'': 1.49\% (Gemma), 2.59\% (Llama)
  \item ``metaphor'': 0.84\% (Gemma), 0.93\% (Llama)
\end{itemize}

\paragraph{Remedy: orthogonalization.}
We project the vector onto the subspace orthogonal to all problematic token directions:
\begin{equation}
    \mathbf{v}_{clean} = \mathbf{v}_{analogy} - \sum_{t \in T} \text{proj}_{\mathbf{d}_t} \mathbf{v}_{analogy}
\end{equation}
where $T$ is the set of problematic tokens and $\mathbf{d}_t$ is the unembedding direction for token $t$. This approach differs from \citet{arditi2024refusal}, who use orthogonalization to permanently ablate a direction from model weights; here we apply it for \emph{vector refinement} at inference time, preserving the underlying capability while removing the lexical artifact.

\paragraph{Effect on vector norm.}
The operation reduces vector magnitude by only 1--5\% across layers (3--5\% in late layers where token directions align most with output), confirming that the analogy direction is predominantly semantic (90--95\% of the norm) rather than a lexical trigger.

\paragraph{Quantitative results.}
At $\alpha=2.0$, the original vector causes severe collapse: ``Imagine'' count rises from 0--2 (baseline) to 37--40 for Gemma and 0--1 to 4 for Llama, while harmonic mean drops from 3.0 to 1.4 and 2.8 to 2.6 respectively. The clean vector reduces ``Imagine'' occurrences by 92\% for Gemma (to 3--6) and 75\% for Llama (to 1), with corresponding recovery in harmonic mean. At validated operating coefficients ($\alpha=0.5$--$1.0$), both dirty and clean vectors perform similarly---the clean vector is primarily a guard against high-coefficient failure, not a general performance improvement.

\begin{figure}[ht]
\begin{center}
\includegraphics[width=0.85\linewidth]{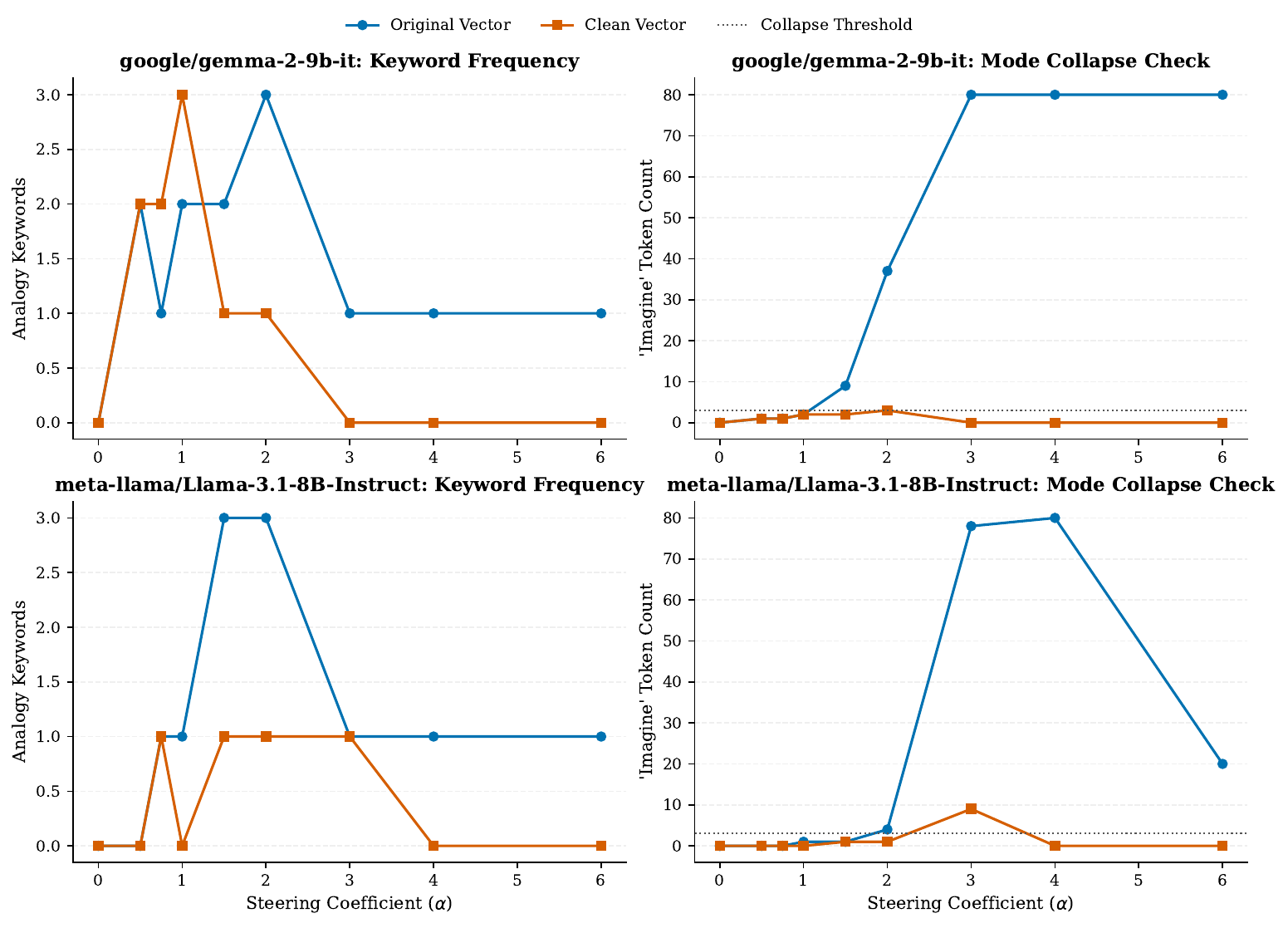}
\end{center}
\caption{Original (Dirty) versus Clean Vector. \textbf{Left}: Both vectors successfully steer the model to produce analogies (keyword count). \textbf{Right}: The original vector causes mode collapse into the ``Imagine'' token at $\alpha \ge 2$, whereas the clean vector suppresses this failure mode by 75--92\%.}
\label{fig:ortho}
\end{figure}

Figure~\ref{fig:ortho} illustrates this disentanglement: the orthogonalized vector preserves steering efficacy (Left) while eliminating mode collapse (Right). The result confirms that the abstract representation of analogy-making is linearly separable from its specific surface-level lexical triggers.

\section{Additional Results}
\label{app:results}

\subsection{Layer Ablation Results}


Table \ref{tab:ablation} shows keyword counts across different layers for Gemma-2-9B-IT at coefficient 0.6. The transition zone (Layer 29, identified via stability analysis) outperforms both early computation layers and late stable layers.

\begin{table}[ht]
\caption{Layer Ablation Results (Gemma-2-9B-IT, $\alpha=0.6$). Keyword counts show steering effectiveness across layers.}
\label{tab:ablation}
\begin{center}
\begin{tabular}{lcc}
\toprule
\bf Layer & \bf Keywords & \bf In Stable Region \\
\midrule
28 & 2 & No \\
\textbf{29} & \textbf{3} & \textbf{No (Transition)} \\
30--31 & 2 & No \\
32--41 & 1--2 & Yes \\
\bottomrule
\end{tabular}
\end{center}
\end{table}

\subsection{Exploratory: Cosine Stability Analysis}
As a complementary exploratory analysis, we examined inter-layer cosine similarity of $\mathbf{v}_{analogy}$ to identify computation-to-propagation transitions. All main results use the empirically selected layers, \textit{not} the stability heuristic.

\begin{itemize}
    \item \textbf{Stable regions exist in late layers} for both models (Layers 32--41 for Gemma, 26--31 for Llama), with cosine similarity $>0.95$ between adjacent layers
    \item \textbf{Optimal steering at transition zones} (Layer 29 for Gemma, 24 for Llama), where the concept transitions from computation to propagation
    \item \textbf{Low coefficients effective} (0.5--1.0) without mode collapse, suggesting the mechanism is sensitive and doesn't require strong interventions
    \item \textbf{Orthogonalization reduces mode collapse} by 75--92\% in both models, confirming the approach generalizes across architectures
    \item \textbf{Model-specific responsiveness:} Gemma shows stronger steering effect (3.0 to 4.0, 33\% improvement) compared to Llama (2.8 to 3.2, 14\% improvement), but both demonstrate clear improvement over baseline
    \item \textbf{Token entanglement patterns:} ``Imagine'' accounts for 10.27\% of vector in Gemma and 7.68\% in Llama, with similar token rankings across models
\end{itemize}
This suggests analogy-making uses a general mechanism that transcends architecture, though optimal coefficients and layer selection may require model-specific tuning.

\subsection{Mode Collapse Analysis}
At $\alpha=2.0$, the original (dirty) vector causes severe mode collapse, reflected in both evaluation metrics:
\begin{itemize}
    \item \textbf{Gemma:} ``Imagine'' count increases from 0--2 (baseline) to 37--40 (collapse), while quality scores drop from 3.0 to 1.4, representing complete degradation
    \item \textbf{Llama:} ``Imagine'' count increases from 0--1 (baseline) to 4 (collapse), with quality scores dropping from 2.8 to 2.6, showing more gradual degradation
\end{itemize}
The clean vector reduces this to 3--6 for Gemma (92\% reduction) and 1 for Llama (75\% reduction), demonstrating that orthogonalization successfully disentangles the semantic concept from lexical triggers. At $\alpha=2.0$, Llama's clean vector achieved a quality score of 3.0 compared to the original vector's 2.6, demonstrating that orthogonalization not only prevents collapse but can actually improve performance at intermediate coefficients.

At validated coefficients (0.5--1.0), both dirty and clean vectors work similarly (1--3 keywords, 0--2 ``imagine'' counts, quality scores 3.8--4.0), confirming the clean vector preserves the analogy concept. At transition coefficients (1.5--2.0), the clean vector's advantage becomes clear: Gemma shows $\ge$90\% reduction in ``Imagine'' counts (37 to 3, 9 to 2), while Llama shows 75\% reduction (4 to 1), with corresponding improvements in quality scores.

At very high coefficients ($\alpha \ge 3.0$), the clean vector avoids ``Imagine'' collapse but may collapse into other tokens (``Okay,'' ``picture''), suggesting additional entanglement may exist with other analogy-related tokens, or that the vector magnitude itself becomes problematic at extreme values. However, even at these extreme coefficients, the clean vector shows more graceful degradation than the original vector, with quality scores remaining above 1.0 compared to the original vector's complete collapse to 1.0.

\subsection{Base-Instruct Vector Similarity: 
            Geometric Evidence for the Gradient}
\label{app:cosine_sim}

To corroborate the behavioral gradient with geometric evidence, 
we computed cosine similarity between analogy vectors from each 
matched base-instruct pair. Table~\ref{tab:cosine_sim} reports 
two metrics: \textbf{Max}, the global maximum across all layer 
combinations, and \textbf{Same-Layer}, the similarity at the 
IT model's empirically best layer for both models.

\begin{table}[ht]
\caption{Base-instruct cosine similarity. \textbf{Max}: global 
maximum across all $(\ell_{\text{base}}, \ell_{\text{IT}})$ 
pairs. \textbf{Same-Layer}: both models evaluated at 
$\ell^*_{\text{IT}}$.}
\label{tab:cosine_sim}
\begin{center}
\small
\begin{tabular}{lccc}
\toprule
\bf Family & \bf Max & \bf Same-Layer & \bf Interpretation \\
\midrule
Gemma-2 (2B/9B/27B)   & 0.37--0.66 & 0.31--0.55 & Reorganized \\
Llama-3.1 (8B)        & 0.30 & 0.23 & Different \\
Qwen-2.5 (7B/14B/32B) & 0.94--0.97 & 0.17--0.46 & Preserved \\
\bottomrule
\end{tabular}
\end{center}
\end{table}

Qwen base-IT pairs show near-identity peaks 
(cos sim $>$ 0.93), while Gemma pairs show substantially 
weaker alignment ($<$ 0.66) even at the most favorable layer 
pairing. This quantifies—independent of behavioral metrics—the 
geometric transformation IT performs on a specific feature 
direction, a finding not predictable from pretraining 
documentation alone. Combined with the behavioral decomposition 
(for Gemma-2-9B-IT, steering captures $\sim$55\% of prompting's 
effect: literal HM$=$3.22 $\to$ steered 3.94 $\to$ prompted 
4.54), this clarifies steering's mechanistic role: 
\textbf{for Qwen, steering reinforces an intact pathway; for 
Gemma, it amplifies a pathway IT constructed}. IT builds the 
wiring; the analogy vector provides targeted amplification. The 
gradient claim rests on peak similarity magnitude, not layer 
location.

Table~\ref{tab:cosine_sim_full} reports per-model cosine 
similarity between analogy vectors at three reference points:
\textbf{Max} (global maximum across all layer pairs), 
\textbf{At Optimal} (base at $\ell^*_{\text{base}}$, IT at 
$\ell^*_{\text{IT}}$—from Table~\ref{tab:layer_selection}), 
and \textbf{Same-Layer} (both at $\ell^*_{\text{IT}}$).

\begin{table}[ht]
\caption{Full base-IT cosine similarity results (per model).}
\label{tab:cosine_sim_full}
\begin{center}
\small
\begin{tabular}{lccc}
\toprule
\bf Pair & \bf Max & \bf At Optimal & \bf Same-Layer \\
\midrule
Gemma-2-2B    & 0.574 & 0.081  & 0.524 \\
Gemma-2-9B    & 0.373 & 0.099  & 0.310 \\
Gemma-2-27B   & 0.655 & -0.107 & 0.551 \\
Llama-3.1-8B  & 0.300 & 0.178  & 0.234 \\
Qwen-2.5-7B   & 0.939 & 0.462  & 0.462 \\
Qwen-2.5-14B  & 0.954 & 0.286  & 0.395 \\
Qwen-2.5-32B  & 0.965 & 0.152  & 0.167 \\
\bottomrule
\end{tabular}
\end{center}
\end{table}

\section{Example Generations}
\label{app:examples}

This appendix is organised in two parts.
\textbf{Part~A} shows the motivation for orthogonalization: the same TCP/IP prompt steered with the \emph{raw} (dirty) vector illustrates mode collapse at $\alpha{=}2.0$, confirming that all test-set results in the main paper use the \emph{clean} (orthogonalized) vector.
\textbf{Part~B} gives one example per instruction-tuned model from the 300-prompt held-out test set (seed~42), spanning seven distinct knowledge domains, all steered with the clean vector at the empirically best layer and reported $\alpha$.
Each box shows three conditions on the same literal prompt:
\textbf{Prompting Baseline} (analogy-eliciting prompt, no steering),
\textbf{Unsteered Literal} ($\alpha=0$), and \textbf{Steered Clean}.
Outputs are truncated at $\sim$350 characters for space.


\subsection*{Part A\quad Motivation for Orthogonalization: Mode Collapse with Raw Vector}

The example below (Gemma-2-9B-IT, TCP/IP handshakes) uses the \textbf{raw} analogy vector \emph{before} orthogonalization.
At $\alpha{=}1.0$ steering already works; at $\alpha{=}2.0$ the ``\texttt{Imagine}'' token dominates ($\sim$10\% of vector norm) and the model collapses into a repetition loop.
All seven per-model examples in Part~B use the clean vector and show no such collapse.

\begin{tcolorbox}[colback=gray!10,colframe=gray!50,breakable,
  title={\textbf{Concept:} TCP/IP handshakes \quad \textbf{Model:} Gemma-2-9B-IT \quad \textbf{Vector:} raw (not orthogonalized)},
  fonttitle=\bfseries]

\textbf{Unsteered Literal} ($\alpha=0$):
\begin{tcolorbox}[colback=white,colframe=gray!30,boxrule=0.5pt,left=3pt,right=3pt,top=2pt,bottom=2pt]
\small\ttfamily
The TCP/IP handshake is a three-way process that establishes a reliable connection between two devices. It ensures both devices are ready to communicate and understand the parameters of the connection.
1.~SYN: The client sends a SYN packet containing a sequence number.
2.~SYN-ACK: The server acknowledges and sends its own sequence number.
3.~ACK: The client acknowledges the server's sequence number. The connection is established...
\end{tcolorbox}

\vspace{0.15cm}
\textbf{Steered (raw vector, $\alpha=1.0$)} --- effective steering:
\begin{tcolorbox}[colback=blue!8,colframe=blue!40,boxrule=0.5pt,left=3pt,right=3pt,top=2pt,bottom=2pt]
\small\ttfamily
Imagine you want to have a conversation with someone across the room. You wouldn't just start talking---you'd knock first to see if they're ready. That's how TCP/IP handshakes work: like knocking on a door before sending data.
Think of it as a three-step dance:
1.~Imagine you (the client) want to talk to a server. You send a ``knock'' (SYN packet) saying ``I want to connect.''
2.~The server replies ``I hear you, and I'm ready'' (SYN-ACK).
3.~You confirm ``Great, let's go'' (ACK). The connection is established...
\end{tcolorbox}

\vspace{0.15cm}
\textbf{Steered (raw vector, $\alpha=2.0$)} --- mode collapse:
\begin{tcolorbox}[colback=red!8,colframe=red!50,boxrule=0.5pt,left=3pt,right=3pt,top=2pt,bottom=2pt]
\small\ttfamily
Imagine you want to imagine imagine imagine imagine imagine imagine think of it like imagine imagine imagine imagine okay let's imagine you want to imagine imagine imagine imagine okay imagine imagine imagine imagine imagine imagine imagine imagine imagine imagine imagine imagine...
\end{tcolorbox}
\textit{\small The ``\texttt{Imagine}'' token embedding contributes $\sim$10\% of the raw vector's norm.
Orthogonalizing against this token direction reduces the magnitude by only 1--5\% while eliminating collapse entirely (see Appendix~\ref{app:ortho}).}

\end{tcolorbox}

\subsection*{Part B\quad Per-Model Examples (Clean Vector)}

\subsection{Gemma-2-2B-IT \texorpdfstring{\textnormal{(Physics --- Sound waves at the speed of sound, layer~9, $\alpha$=1.0)}}{(Physics --- Sound waves at the speed of sound, layer 9, alpha=1.0)}}

\begin{tcolorbox}[colback=gray!10, colframe=gray!50, breakable,
  title={\textbf{Prompt (Literal/Steered):} ``Provide a precise, factual explanation of hearing a sound wave while traveling at the speed of sound.''},
  fonttitle=\bfseries]

\textbf{Prompting Baseline} (analogy-eliciting prompt):
\begin{tcolorbox}[colback=gray!5,colframe=gray!40,boxrule=0.5pt,left=3pt,right=3pt,top=2pt,bottom=2pt]
\small\ttfamily
Imagine a pebble dropped into a still pond. The ripple it creates spreads outward, growing larger as it travels.
Now imagine that ripple is a sound wave. The pebble is the sound source; the pond is the air; the ripples are the sound waves; the speed of sound is like the speed of the ripple. If you travel alongside the ripple at exactly its speed, you'd always be at the edge of the wave---you'd never hear it pass you...
\end{tcolorbox}

\vspace{0.15cm}
\textbf{Unsteered Literal} ($\alpha=0$):
\begin{tcolorbox}[colback=white,colframe=gray!30,boxrule=0.5pt,left=3pt,right=3pt,top=2pt,bottom=2pt]
\small\ttfamily
Sound is a mechanical wave requiring a medium (air, water, solids) to travel. When a sound source vibrates, it creates compressions and rarefactions in the surrounding medium. Compression: areas where molecules are closer together. Rarefaction: areas where the medium is less dense. The speed of sound in air at room temperature is approximately 343~m/s...
\end{tcolorbox}

\vspace{0.15cm}
\textbf{Steered} ($\alpha=1.0$, layer~9):
\begin{tcolorbox}[colback=blue!8,colframe=blue!40,boxrule=0.5pt,left=3pt,right=3pt,top=2pt,bottom=2pt]
\small\ttfamily
Imagine you're on a train traveling at the speed of sound. The sound wave is the train itself---moving at the speed of sound, it's like a giant invisible wave of energy. The sound wave is a ripple in a pond: spreading out from the train's movement, but small and contained. The sound travels through the air just like the ripple, yet it doesn't form a large wave, just a precise, directed disturbance traveling at exactly your speed...
\end{tcolorbox}
\end{tcolorbox}

\subsection{Gemma-2-9B-IT \texorpdfstring{\textnormal{(Biology --- Topical pain relievers on internal pain, layer~18, $\alpha$=1.0)}}{(Biology --- Topical pain relievers on internal pain, layer 18, alpha=1.0)}}

\begin{tcolorbox}[colback=gray!10,colframe=gray!50,breakable,
  title={\textbf{Prompt (Literal/Steered):} ``Explain the mechanism of topical pain relievers on internal pain using accurate technical terminology.''},
  fonttitle=\bfseries]

\textbf{Prompting Baseline}:
\begin{tcolorbox}[colback=gray!5,colframe=gray!40,boxrule=0.5pt,left=3pt,right=3pt,top=2pt,bottom=2pt]
\small\ttfamily
It's like trying to put out a fire in your stomach with a fire extinguisher designed for a kitchen fire. Topical pain relievers are like that kitchen extinguisher: they work by numbing the application site, blocking pain signals locally via lidocaine or menthol. Internal pain is like the fire in your stomach---it originates from organs and tissues well beyond the skin's reach. The fire extinguisher simply cannot reach it...
\end{tcolorbox}

\vspace{0.15cm}
\textbf{Unsteered Literal} ($\alpha=0$):
\begin{tcolorbox}[colback=white,colframe=gray!30,boxrule=0.5pt,left=3pt,right=3pt,top=2pt,bottom=2pt]
\small\ttfamily
Topical pain relievers, when applied to the skin, do \textit{not} directly reach internal pain sources. The skin acts as a formidable barrier preventing most substances from penetrating deeply. While some topical medications are absorbed transdermally, the amount reaching the bloodstream is typically insufficient to impact visceral or deep musculoskeletal pain. Topical NSAIDs (e.g., diclofenac) may achieve local anti-inflammatory concentrations in subcutaneous tissue but not in deep internal structures...
\end{tcolorbox}

\vspace{0.15cm}
\textbf{Steered} ($\alpha=1.0$, layer~18):
\begin{tcolorbox}[colback=blue!8,colframe=blue!40,boxrule=0.5pt,left=3pt,right=3pt,top=2pt,bottom=2pt]
\small\ttfamily
Think of your skin as a barrier. Topical pain relievers like creams or patches work by blocking pain signals at the ``gatekeeper'' level of your spinal cord. Internal pain is like a fire in your stomach---those signals travel directly to the brain, bypassing the gatekeeper. Imagine a patch on your knee: it can numb the nearby nerves but cannot send its active ingredient through layers of muscle and tissue to reach a distant organ...
\end{tcolorbox}
\end{tcolorbox}

\subsection{Gemma-2-27B-IT \texorpdfstring{\textnormal{(Economics --- Credit card cancellation and credit score, layer~16, $\alpha$=1.0)}}{(Economics --- Credit card cancellation and credit score, layer 16, alpha=1.0)}}

\begin{tcolorbox}[colback=gray!10,colframe=gray!50,breakable,
  title={\textbf{Prompt (Literal/Steered):} ``Explain credit card cancellation affecting credit score using accurate technical terminology.''},
  fonttitle=\bfseries]

\textbf{Prompting Baseline}:
\begin{tcolorbox}[colback=gray!5,colframe=gray!40,boxrule=0.5pt,left=3pt,right=3pt,top=2pt,bottom=2pt]
\small\ttfamily
Imagine your credit score is like a plant. Credit cards are watering cans---using them responsibly helps your plant grow strong. Cancelling a card removes one watering can, reducing your ability to water evenly. Your credit utilization ratio is the ratio of water used to total watering capacity; fewer cans means the same amount of water represents a higher fraction. Length of credit history is like the age of the plant---older plants are generally stronger and more resilient...
\end{tcolorbox}

\vspace{0.15cm}
\textbf{Unsteered Literal} ($\alpha=0$):
\begin{tcolorbox}[colback=white,colframe=gray!30,boxrule=0.5pt,left=3pt,right=3pt,top=2pt,bottom=2pt]
\small\ttfamily
Cancelling a credit card affects your credit score through two primary mechanisms: (1) \textbf{Credit utilization ratio}: calculated as total balances divided by total credit limits. Cancelling a card reduces your total credit limit, potentially increasing utilization---especially if you carry balances on remaining cards. FICO models weight utilization at $\sim$30\% of the score. (2) \textbf{Credit history length}: the average age of open accounts may decrease if the cancelled card is among your oldest...
\end{tcolorbox}

\vspace{0.15cm}
\textbf{Steered} ($\alpha=1.0$, layer~16):
\begin{tcolorbox}[colback=blue!8,colframe=blue!40,boxrule=0.5pt,left=3pt,right=3pt,top=2pt,bottom=2pt]
\small\ttfamily
Think of your credit report like a financial resume. Lenders use it to assess your trustworthiness. Your credit utilization ratio is like a pie chart of available credit: each card is a slice. Cancel a card and you're shrinking the pie---the same balance now represents a larger fraction, signalling higher risk. Your credit history length is like the years of experience on your resume: removing an old card can shorten it, weakening the overall picture you present to lenders...
\end{tcolorbox}
\end{tcolorbox}

\subsection{Llama-3.1-8B-IT \texorpdfstring{\textnormal{(Engineering --- Why guns can discharge when dropped, layer~11, $\alpha$=1.5)}}{(Engineering --- Why guns can discharge when dropped, layer 11, alpha=1.5)}}

\begin{tcolorbox}[colback=gray!10,colframe=gray!50,breakable,
  title={\textbf{Prompt (Literal/Steered):} ``Explain guns going off when dropped using accurate technical terminology.''},
  fonttitle=\bfseries]

\textbf{Prompting Baseline}:
\begin{tcolorbox}[colback=gray!5,colframe=gray!40,boxrule=0.5pt,left=3pt,right=3pt,top=2pt,bottom=2pt]
\small\ttfamily
Imagine a coiled spring. Held in your hand, the spring is still and its energy stored. If you drop it, the spring uncoils and releases that energy. A gun's trigger mechanism is similar: when dropped onto a hard surface, the impact can release the spring-loaded striker or firing pin. This is not a design goal---modern firearms incorporate drop-safety mechanisms to prevent exactly this. An older or worn firearm may lack adequate drop safety, allowing inertia to actuate the trigger...
\end{tcolorbox}

\vspace{0.15cm}
\textbf{Unsteered Literal} ($\alpha=0$):
\begin{tcolorbox}[colback=white,colframe=gray!30,boxrule=0.5pt,left=3pt,right=3pt,top=2pt,bottom=2pt]
\small\ttfamily
A ``negligent discharge'' or ``accidental discharge'' can occur when a dropped firearm activates its trigger mechanism. Upon impact with a hard surface, the inertia of the firing pin or striker may overcome the sear engagement, driving the pin forward into the primer. Single-action revolvers and older pistol designs lacking transfer-bar or firing-pin-block safeties are particularly susceptible. Modern double-action firearms incorporate drop-safety mechanisms to prevent inertial discharge...
\end{tcolorbox}

\vspace{0.15cm}
\textbf{Steered} ($\alpha=1.5$, layer~11):
\begin{tcolorbox}[colback=blue!8,colframe=blue!40,boxrule=0.5pt,left=3pt,right=3pt,top=2pt,bottom=2pt]
\small\ttfamily
Think of the gun as a complex machine with interconnected parts, like a set of dominoes. Pulling the trigger is like tipping the first domino---it sets off a chain reaction. When you drop the gun, it's like bumping the table the dominoes are on: the impact can tip that first domino without your intent. The firing pin is the critical domino---if inertia drives it forward past the sear, it strikes the primer and the chain completes. Drop-safety mechanisms are the equivalent of gluing certain dominoes in place...
\end{tcolorbox}
\end{tcolorbox}

\subsection{Qwen-2.5-7B-IT \texorpdfstring{\textnormal{(History/Society --- The experience of d\'ej\`a vu, layer~13, $\alpha$=2.0)}}{(History/Society --- The experience of deja vu, layer 13, alpha=2.0)}}

\begin{tcolorbox}[colback=gray!10,colframe=gray!50,breakable,
  title={\textbf{Prompt (Literal/Steered):} ``Provide a precise, factual explanation of the experience of d\'ej\`a vu in the mind.''},
  fonttitle=\bfseries]

\textbf{Prompting Baseline}:
\begin{tcolorbox}[colback=gray!5,colframe=gray!40,boxrule=0.5pt,left=3pt,right=3pt,top=2pt,bottom=2pt]
\small\ttfamily
D\'ej\`a vu can be analogized to walking into a room that looks exactly like a place you've visited many times---but you've never actually been there. Your brain is like a vast library: when you encounter a new situation, it rapidly searches for matching entries. D\'ej\`a vu occurs when the brain's pattern-matching system finds a strong but erroneous match, generating a feeling of familiarity without a corresponding memory. The hippocampus, your brain's indexing librarian, may momentarily misfile a new experience...
\end{tcolorbox}

\vspace{0.15cm}
\textbf{Unsteered Literal} ($\alpha=0$):
\begin{tcolorbox}[colback=white,colframe=gray!30,boxrule=0.5pt,left=3pt,right=3pt,top=2pt,bottom=2pt]
\small\ttfamily
D\'ej\`a vu (French: ``already seen'') is a phenomenon in which a person feels strong familiarity with a current situation despite no verifiable prior exposure. It is associated with the hippocampus and adjacent medial temporal lobe structures. Leading hypotheses include: (1) a mismatch between perceptual input and memory retrieval signals; (2) transient disruption of the familiarity/recollection distinction; (3) propagation of an epileptiform discharge through memory circuits. Prevalence is $\sim$60\% of the general population...
\end{tcolorbox}

\vspace{0.15cm}
\textbf{Steered} ($\alpha=2.0$, layer~13):
\begin{tcolorbox}[colback=blue!8,colframe=blue!40,boxrule=0.5pt,left=3pt,right=3pt,top=2pt,bottom=2pt]
\small\ttfamily
D\'ej\`a vu is like a moment when the mind is transported back in time, as if you've experienced something before even though you haven't. Imagine walking down a street and feeling as if you've been there before---your brain is trying to make sense of this new situation by comparing it to past experiences. It activates a network of neurons associated with familiar patterns. When the match is close but not exact, the brain generates that eerie ``I've been here'' sensation even in the absence of an actual prior memory...
\end{tcolorbox}
\end{tcolorbox}

\subsection{Qwen-2.5-14B-IT \texorpdfstring{\textnormal{(Chemistry --- pH measurement in liquids, layer~24, $\alpha$=2.0)}}{(Chemistry --- pH measurement in liquids, layer 24, alpha=2.0)}}

\begin{tcolorbox}[colback=gray!10,colframe=gray!50,breakable,
  title={\textbf{Prompt (Literal/Steered):} ``Describe the mechanics behind pH measurement in liquids in literal terms.''},
  fonttitle=\bfseries]

\textbf{Prompting Baseline}:
\begin{tcolorbox}[colback=gray!5,colframe=gray!40,boxrule=0.5pt,left=3pt,right=3pt,top=2pt,bottom=2pt]
\small\ttfamily
Imagine a seesaw in a playground. One side represents acidic substances, the other basic. The fulcrum is neutral---pure water at pH~7. Adding an acid tips the seesaw toward the acid side (more H\textsuperscript{+} ions); adding a base tips it the other way (more OH\textsuperscript{-}). A pH meter is like a precise scale that measures exactly how far from center the seesaw has tipped, reading the concentration of H\textsuperscript{+} ions on a logarithmic scale from 0 to 14...
\end{tcolorbox}

\vspace{0.15cm}
\textbf{Unsteered Literal} ($\alpha=0$):
\begin{tcolorbox}[colback=white,colframe=gray!30,boxrule=0.5pt,left=3pt,right=3pt,top=2pt,bottom=2pt]
\small\ttfamily
pH is defined as $-\log_{10}[\text{H}^+]$, where $[\text{H}^+]$ is the molar concentration of hydrogen ions. Measurement methods include: (1) \textbf{pH meter}: a glass electrode sensitive to H\textsuperscript{+} generates a potential difference measured against a reference electrode; the Nernst equation relates voltage to ion concentration. (2) \textbf{pH indicator strips}: dyes that undergo colorimetric changes at known pH values allow visual estimation. Calibration with buffer solutions of known pH is required for accuracy...
\end{tcolorbox}

\vspace{0.15cm}
\textbf{Steered} ($\alpha=2.0$, layer~24):
\begin{tcolorbox}[colback=blue!8,colframe=blue!40,boxrule=0.5pt,left=3pt,right=3pt,top=2pt,bottom=2pt]
\small\ttfamily
Imagine a swimming pool and you want to know how much chlorine keeps it safe. pH measurement is similar---instead of chlorine, you're checking how many hydrogen ions (H\textsuperscript{+}) float in the water. A pH meter is like a sensitive thermometer for ion concentration: dip it in, and it reads the voltage generated by those ions against a reference, converting the reading to the familiar 0--14 scale via the Nernst equation. Fewer H\textsuperscript{+} ions means a higher pH reading---the pool is more basic, the signal weaker...
\end{tcolorbox}
\end{tcolorbox}

\subsection{Qwen-2.5-32B-IT \texorpdfstring{\textnormal{(Earth/Environment --- Saltwater formation and distribution, layer~25, $\alpha$=2.0)}}{(Earth/Environment --- Saltwater formation and distribution, layer 25, alpha=2.0)}}

\begin{tcolorbox}[colback=gray!10,colframe=gray!50,breakable,
  title={\textbf{Prompt (Literal/Steered):} ``Describe the mechanics behind saltwater formation and distribution on Earth in literal terms.''},
  fonttitle=\bfseries]

\textbf{Prompting Baseline}:
\begin{tcolorbox}[colback=gray!5,colframe=gray!40,boxrule=0.5pt,left=3pt,right=3pt,top=2pt,bottom=2pt]
\small\ttfamily
Imagine the Earth as a giant kitchen and the oceans as a huge salty soup pot. The salts and minerals are the spices, arriving from rocks on land (weathering), underwater volcanoes (hydrothermal vents), and atmospheric deposition. Evaporation is like ladling water out of the pot---it removes only pure water, leaving the spices behind and making the remaining soup progressively saltier. Ocean currents are like stirring the pot: they distribute the salinity globally so no region is far too salty or far too fresh...
\end{tcolorbox}

\vspace{0.15cm}
\textbf{Unsteered Literal} ($\alpha=0$):
\begin{tcolorbox}[colback=white,colframe=gray!30,boxrule=0.5pt,left=3pt,right=3pt,top=2pt,bottom=2pt]
\small\ttfamily
Saltwater formation results from chemical weathering of continental rocks, hydrothermal vent activity at mid-ocean ridges, and evaporite dissolution. Sodium (Na\textsuperscript{+}) and chloride (Cl\textsuperscript{-}) ions dominate oceanic salinity ($\sim$3.5\% by mass). The hydrological cycle concentrates salts: precipitation is nearly ion-free, while evaporation removes only water, progressively enriching the residual solution. Thermohaline circulation distributes saline water globally, with dense, cold, saline deep water forming primarily in polar regions and driving the global conveyor belt...
\end{tcolorbox}

\vspace{0.15cm}
\textbf{Steered} ($\alpha=2.0$, layer~25):
\begin{tcolorbox}[colback=blue!8,colframe=blue!40,boxrule=0.5pt,left=3pt, right=3pt,top=2pt,bottom=2pt]
\small\ttfamily
Imagine the Earth as a giant kitchen: the oceans are a big pot of water, and the continental rocks are like blocks of salt dissolving into it over geological time. Rain falling on rock is like pouring water over a salt block---it slowly dissolves minerals and carries them to the sea. Evaporation removes only the water, like the steam rising from a pot, leaving the dissolved minerals behind. Ocean currents are the kitchen's stirring spoon, distributing salinity so that no region stagnates into a brine pool while another stays nearly fresh...
\end{tcolorbox}
\end{tcolorbox}

\end{document}